\documentclass[sigconf]{acmart}

\usepackage{booktabs} 
\usepackage{amsmath}
\usepackage{algorithm, algpseudocode}
\usepackage{multirow}
\usepackage{comment}
\usepackage{graphicx}
\usepackage{subcaption}
\setcopyright{acmcopyright}

\usepackage{xcolor}

\begin{document}
\title[Jointly Learning Word Embeddings and Latent Topics]{Jointly
  Learning Word Embeddings and Latent Topics}
\thanks{The work described in this paper is substantially supported by grants 
from 
the Research Grant Council of the Hong Kong Special Administrative 
Region, China 
(Project Code: 14203414) and the Microsoft Research Asia Urban Informatics 
Grant FY14-RES-Sponsor-057. Steven Schockaert and Shoaib Jameel are supported by ERC Starting Grant 637277.
}

\author{Bei Shi}
\affiliation{
  \department{Department of Systems Engineering and Engineering Management}
  \institution{The Chinese University of Hong Kong}
  \city{Hong Kong}
}
\email{bshi@se.cuhk.edu.hk}

\author{Wai Lam}
\affiliation{
  \department{Department of Systems Engineering and Engineering Management}  
  \institution{The Chinese University of Hong Kong}
  \city{Hong Kong}
}
\email{wlam@se.cuhk.edu.hk}

\author{Shoaib Jameel}
\affiliation{
  \department{School of Computer Science and Informatics}
  \institution{Cardiff University}
  \city{Cardiff}
  \country{UK}
}
\email{JameelS1@cardiff.ac.uk}

\author{Steven Schockaert}
\affiliation{
  \department{School of Computer Science and Informatics}
  \institution{Cardiff University}
  \city{Cardiff}
  \country{UK}
}
\email{SchockaertS1@cardiff.ac.uk}

\author{Kwun Ping Lai}
\affiliation{
  \department{Department of Systems Engineering and Engineering Management}  
  \institution{The Chinese University of Hong Kong}
  \city{Hong Kong}
}
\email{kplai@se.cuhk.edu.hk}
\renewcommand{\shortauthors}{B. Shi et. al.}
\newcommand{\tabincell}[2]{\begin{tabular}{@{}#1@{}}#2\end{tabular}}
\fancyhead{}

\begin{abstract}
  Word embedding models such as Skip-gram learn a vector-space
  representation for each word, based on the local word collocation
  patterns that are observed in a text corpus. Latent topic models, on
  the other hand, take a more global view, looking at the word
  distributions across the corpus to assign a topic to each word
  occurrence. These two paradigms are complementary in how they
  represent the meaning of word occurrences. While some previous
  works have already looked at using word embeddings for improving the
  quality of latent topics, and conversely, at using latent topics for improving word embeddings, such ``two-step'' methods cannot capture the mutual interaction between the two paradigms. In this paper, we propose STE, a framework which can learn word embeddings and latent topics in a unified manner. STE naturally obtains topic-specific word embeddings, and thus addresses the issue of polysemy. At the same time, it also learns the term distributions of the topics, and the topic distributions of the documents. Our experimental results demonstrate that the STE model can indeed generate useful topic-specific word embeddings and coherent latent topics in an effective and efficient way.
\end{abstract}
\copyrightyear{2017} 
\acmYear{2017} 
\setcopyright{acmcopyright}
\acmConference{SIGIR '17}{}{August 07-11, 2017, Shinjuku, Tokyo, Japan}\acmPrice{15.00}\acmDOI{http://dx.doi.org/10.1145/3077136.3080806}
\acmISBN{978-1-4503-5022-8/17/08}
%
%



\maketitle
\section{Introduction}

Word embeddings, also known as distributed word representations, are a popular way of representing words in Natural Language Processing (NLP) and Information Retrieval (IR) applications \cite{collobert2008unified,
  mikolov2013linguistic, socher2011parsing, turney2010frequency,glove2014}. Essentially, the idea is to represent each word as a vector in a low-dimensional space, in a way which reflects the semantic, and sometimes also syntactic, relationships between the words. One natural requirement is that the vectors of similar words are themselves also similar (e.g.\ in terms of cosine similarity or Euclidean distance). In addition, in some models, several kinds of linear regularities are observed. For example, in Skip-gram \cite{mikolov2013distributed}, one of the most commonly used word embedding models, analogous word pairs tend to form parallelograms in the space, a notable example being vec(``man'') - vec(``king'') $\approx$ vec(``woman'') - vec(``queen''). Most word embedding models rely on statistics about how often each word occurs within a local context window of another word, either implicitly \cite{mikolov2013linguistic} or explicitly \cite{turney2010frequency,glove2014}.


Topic models, such as Latent Dirichlet Allocation (LDA)~\cite{blei2003latent}, assign a discrete topic to each word occurrence in a corpus. These topics can be seen as groups of semantically related words. In this sense, like word embeddings, topic models can be viewed as models for capturing the meaning of the words in a corpus. However, there are several key differences between word embeddings and topic models, which make them complementary to each other. First, word embeddings are continuous representations, whereas topic assignments are discrete. Second, word embeddings are learned from local context windows, whereas topic models take a more global view, in the sense that the topic which is assigned to a given word occurrence (in the case of LDA) equally depends on all the other words that appear in the same document. Several researchers have already exploited this complementary representation between word embeddings and topic models.


 On the one hand, topic models can be used to improve word embeddings, by addressing the problem of polysemous words. Standard word embedding models essentially ignore ambiguity, meaning that the representation of a word such as ``apple'' is essentially a weighted average of a vector that would intuitively represent the fruit and a vector that would intuitively represent the company. A natural solution, studied in Liu et al.~\cite{liu2015topical}, is to learn different word embeddings for each word-topic combination. In particular, they propose a model called Topical Word Embeddings (TWE), which first employs the standard LDA model to obtain word-topic assignments. Regarding each topic as a pseudo-word, they then learn embeddings for both words and topics. Finally, a given word-topic combination is represented as the concatenation of the word vector and the topic vector.


On the other hand, word embeddings can also be used to improve topic models. For example, Nguyen et al.~\cite{nguyen2015improving} suggest to model topics as mixtures of the usual Dirichlet multinomial model and a word embedding component. It is shown that the top words associated with the resulting topics are semantically more coherent. Word embeddings can also be used to help with identifying topics for short texts or small collections. For example, Li et al.~\cite{li2016topic} propose a model which can promote semantically related words, identified by the word embedding, by using the generalized Polya urn model during the sampling process for a given topic. In this way, the external knowledge about semantic relatedness that is captured by the word embedding is exploited to alleviate sparsity problems.


While combining word embeddings and topic models is clearly beneficial, existing approaches merely apply a pipeline approach, where either a standard word embedding is used to improve a topic model, or a standard topic model is used to learn better word embeddings. Such two-step approaches cannot capture the mutual reinforcement between the two types of models. For example, knowing that ``apple'' occurs in two topics can help us to learn better word embeddings, which can in turn help us to learn better topic assignments, etc. The research question which we address in this paper is whether a unified framework, in which topic assignments and word embeddings are jointly learned, can yield better results than the existing two-step approaches. 
The unified framework we propose, named STE, can learn different topic-specific word embeddings, and thus addresses the problem of polysemy, while at the same time generating the term distributions of topics and topic distributions of documents. Our hypothesis is that this will lead both to more meaningful embeddings and more coherent topics, compared to the current state-of-the-art.

From a technical point of view, there are two challenges that need to be addressed. The first challenge concerns the representation of topics with embedding vectors, and the mechanism by which words are generated from such topics. Clearly, the commonly used multinomial distribution is inappropriate in our setting. The second challenge is to obtain the embedding vectors efficiently. Because of the huge amount of parameters, the traditional Skip-gram model exploits the Hierarchical Softmax Tree~\cite{morin2005hierarchical} or Negative Sampling method to maximize the likelihood. When latent topics are considered, however, this alone does not lead to a sufficiently efficient method.


To address the first challenge, we use a generating function that can predict surrounding words, given a target word and its topic. The probability that a given word is generated is based on the inner product of a topic-specific embedding of that word and a topic-specific embedding of the target word. This generating function also allows us to identify the top-ranked words for each topic, which is important for the interpretability of the model. To address the second challenge, we design a scalable EM-Negative Sampling method. This inference method iterates over every skip-gram (i.e.\ each local context window), each time sampling the corresponding negative instances. In the E-step, we evaluate the posterior topic distribution for each skip-gram. In the M-step, we update the topic-specific embeddings and the topic distribution of the documents. We consider two variants of our model, which make different assumptions on the consistency of topics among the word pairs in a skip-gram.

We compare our model with existing hybrid models and perform extensive
experiments on the quality of the word embeddings and latent topics. We also evaluate our performance on the downstream
application of document classification. The experimental results
demonstrate that our model can generate better word embeddings and
more coherent topics than the state-of-the-art models.

\section{Related Work}
In traditional vector space models, individual words are encoded using the so-called one-hot representation, i.e.\ a high-dimensional vector with all zeroes except in one component, corresponding to that word \cite{baeza1999modern}. Such representations suffer from the curse of dimensionality, as there are as many components in these vectors as there are words in the vocabulary. Another important drawback is that semantic relatedness of words cannot be modelled using such representations. To address these shortcomings, Rumelhart et al.~\cite{rumelhart1988learning} propose to use distributed word representation instead, i.e., word embeddings. Several techniques for generating such representations have been investigated. For example, Bengio et al.~\cite{bengio2003neural, bengio2009learning} propose a neural network architecture for this purpose. Later, Mikolov et al.~\cite{mikolov2013distributed} propose two methods that are considerably more efficient, namely Skip-gram and CBOW. This has made it possible to learn word embeddings from large data sets, which has led to the current popularity of word embeddings. Word embedding models have been applied to many tasks, such as named entity recognition~\cite{turian2010word}, word sense disambiguation~\cite{collobert2011natural, iacobacci2016embeddings}, parsing~\cite{roth2016neural}, and information retrieval~\cite{rekabsaz2016enhancing}.


Basic word embedding methods perform poorly for polysemous words such as ``apple'' and ``bank'', as the vectors for such words intuitively correspond to a weighted average of the vectors that would normally be associated with each of the individual senses. Several approaches have been proposed to address this limitation, by learning multiple vectors for each word, one corresponding to each sense \cite{reisinger2010multi, ren2016improving, salehi2015word}. For example, Huang et al.~\cite{huang2012improving} exploit global properties such as term frequency and document frequency to learn multiple embeddings via neural networks. Tian et al.~\cite{tian2014probabilistic} introduce a latent variable to denote the distribution of multiple prototypes for each word in a probabilistic manner. Neelakantan et al.~\cite{neelakantan2015efficient} propose a non-parametric way to evaluate the number of senses for each word. Bartunov et al.~\cite{bartunov2015breaking} also propose a non-parametric Bayesian method to learn the required number of representations. 


Note that our model is different from these models. First, the aforementioned models consider the prototype vectors for each word in isolation, intuitively by clustering the local contexts of each word. This means that these models are limited in how they can model the correlations between the multiple senses of different words. Second, these models do not capture the correlations between the prototypes of words and topics of documents. While there are some word embedding models that do consider topics, to the best of our knowledge no approaches have been studied that exploit the mutual reinforcement between latent topics and word embeddings. For example, Liu et al.~\cite{liu2015topical} concatenate pre-trained topic vectors with the word vectors to represent word prototypes. Building on this idea, Liu et al.~\cite{liu2015learning} combine topic vectors and word vectors via a neural network. 


In traditional topic models, such as LDA~\cite{blei2003latent} and PLSA~\cite{hofmann1999probabilistic}, a document is represented as a multinomial distribution of topics, and the topic assignment of a word only depends on that multinomial distribution. In the Bi-gram Topic Model~\cite{wallach2006topic,wang2007topical}, the topic of a given word additionally depends on the topic of the preceding word.
Our model is related to this bi-gram model, in the sense that the objective functions of both models are based on a similar idea. 


Nguyen et al.~\cite{nguyen2015improving} propose a topic model named LFTM, which generates vectors from a pre-trained word embedding, instead of words. In this way, the model can benefit from the semantic relationships between words to generate better topics. The Gaussian LDA model from Das et al.~\cite{das2015gaussian}, similarly associates with each topic a Gaussian in the word embedding, from which individual word vectors are sampled. Li et al.~\cite{li2016topic} propose a model which can promote semantically related words (given a word embedding) within any given topic. Note that the above models all rely on a pre-trained word embedding. Li et al~\cite{li2016generative} propose a model which learns an embedding link function to connect the word vectors and topics. However, their model mainly focuses on the distributed representation of each topic, instead of words, and generates topics as an abstract vector, thus losing the interpretability of topics.



%
\section{ Model Description}

In this section, we present the details of our model, which we call \textbf{S}kip-gram
\textbf{T}opical word \textbf{E}mbedding (STE).

\subsection{Representing Topics and Embeddings}

 Each word $w$ is associated with an input matrix $U_w$ and an output matrix $V_w$, both of which have dimension $K \times s$, with $K$ the number of topics and $s$ the number of dimensions in the word embedding. The fact that $U_w$ and $V_w$ are matrices, rather than vectors, reflects our modelling assumption that a word $w$ may have a different representation under each topic. 
 
 As in standard topic models, a document will correspond to a probability distribution over topics. In contrast to standard topics models, however, topics in our case are more than probability distributions over words. In particular, for a document $d$ and some central word $w_t$ under consideration, the probability of predicting a surrounding word $w_{t+j}$ depends on the topic of the word $w_t$. For example, suppose that the central word is ``apple''; if its topic relates to technology, words such as ``technology'' might be predicted with high probability, whereas if its topic relates to fruit, words such as ``juice'' might instead be predicted. In particular, we assume that the probability of predicting the word
$w_{t+j}$ given the word $w_t$ under the topic $z$ is evaluated as
follows:
\small
\begin{equation}
    p(w_{t+j}|w_{t}, d) = \sum_{z} p(w_{t+j}|w_t,z) p(z|d)
  \end{equation}
  \normalsize
where the summation is over the set of all $K$ topics, $p(.|d)$ is the topic distribution of the document $d$, and we assume that $j$ is within the window size.

We consider two variants, which differ in how the probability $p(w_{t+j}|w_t,z)$ is evaluated. In the first variant, called STE-Same, we assume that for each skip-gram $\langle w_{t+j}, w_t \rangle$, the words $w_{t+j}$ and $w_t$ belong to the same topic $z$: 
\begin{equation}
  \small
  \label{eq:wz}
    p(w_{t+j}|w_t,z) = \frac{\exp(V_{w_{t+j},z}\cdot
      U_{w_{t},z})}{\sum_{w^{\prime}\in\Lambda}\exp(V_{w^{\prime},z}\cdot
      U_{w_t,z})}
    \normalsize
  \end{equation}
where $\Lambda$ is the vocabulary of the whole corpus. Computing the value $p(w_{t+j}|w_t,z)$ based on Eq.~\ref{eq:wz} is not feasible in practice, given that the computational cost is proportional to the size of $\Lambda$. However, similar as for the standard Skip-gram model, we can rely on negative sampling to address this (see Section \ref{secAlgorithmDesign}).


In the second variant, called STE-Diff, we assume that for each skip-gram $\langle w_{t+j}, w_t \rangle$, the
topic assignment $z_{t+j}$ of word $w_{t+j}$ is independent of the topic assignment $z_t$ of word $w_t$. We then have:
\begin{equation}
  \small
  \begin{split}
  p(w_{t+j}|w_t, d)& =
  \sum_{z_t=1}^{K}\sum_{z_{t+j}=1}^{K}p(w_{t+j}|w_t,z_t, z_{t+j})
  p(z_t, z_{t+j}|d) \\
  & = \sum_{z_t=1}^{K}\sum_{z_{t+j}=1}^{K}p(w_{t+j}|w_t,z_t, z_{t+j})
  p(z_t|d)p(z_{t+j}|d)
\end{split}
\normalsize
\end{equation}
The probability that the word $w_{t+j}$ is generated, given the central word $w_t$ and the topic assignments $z_{t+j}$ and $z_t$ is then evaluated as follows:
\begin{equation}
  \small
  p(w_{t+j}|w_t, z_t, z_{t+j}) = \frac{\exp(V_{w_{t+j},z_{t+j}}\cdot
    U_{w_{t},z_t})}{\sum_{w^{\prime}\in\Lambda}\exp(V_{w^{\prime},z_{t+j}}\cdot
    U_{w_t,z_{t}})}
  \normalsize
\end{equation}


Clearly, both variants have complementary advantages and drawbacks. The STE-Same model will lead to more coherent topics, but it will not allow us to measure the similarity between words across different topics. The STE-Diff model, on the other hand, does allow us to evaluate such inter-topic similarities, but the resulting topics may be less coherent. In practice, we could of course also consider intermediate approaches, where $z_{t+j}=z_j$ is assumed to hold with a high probability, rather than being imposed as a hard constraint. 


\subsection{Algorithm Design}\label{secAlgorithmDesign}

We need an inference method that can learn, given a corpus, the values of the model parameters, i.e.\ the word embeddings $U_{w,z}$ and $V_{w,z}$ corresponding to each topic $z$, as well as the topic distribution $p(z|d)$ for each document $d$. Our inference framework combines the Expectation-Maximization (EM) method with the negative sampling scheme. It is summarized for the STE-Same variant in Algorithm~\ref{alg:em}. The inference method for STE-Diff is analogous. In each iteration of this algorithm, we update the word embeddings and then evaluate the topic distribution $p(z|d)$ of each document. To update the word embeddings, we iterate over each skip-gram, sample several negative instances and then compute the posterior topic distribution for the skip-gram. Then we use the EM algorithm to optimize the log-likelihood of the skip-grams in the document. In the E-step, we use the Bayes rule to evaluate the posterior topic distribution and derive the objective function. In the M-step, we maximize the objective function with the gradient descent method and update the corresponding embeddings $U_w$ and $V_w$.

\begin{algorithm}[t]
  \caption{EM negative sampling for STE-Same}
  \label{alg:em}
  \begin{algorithmic}[1]
    \State Initialize $U$, $V$, $p(z|d)$
    \For {$\textit{out}\_\textit{iter} = 1$ to $\textit{Max}\_\textit{Out}\_\textit{iter}$}
    \For{each document $d$ in $\mathcal{D}$}
    \For{each skip-gram $\langle w_{t+j}, w_t \rangle $ in $d$}
    \State Sample negative instances from the distribution P.
    \State Update $p(w_{t+j}|w_t, z)$, $p(z_k|d, w_{t+j}, w_t)$ by
    Eq.~\ref{eq:sim} and Eq.~\ref{eq:post} respectively.
    \For {$\textit{in}\_\textit{iter} = 1$ to $\textit{Max}\_\textit{In}\_\textit{iter}$}
    \State Update $U$, $V$ using the gradient decent method with
    Eq.~\ref{eq:embedding1} and Eq.~\ref{eq:embedding2}
    \EndFor
    \EndFor
        \State Update $p(z|d)$ using Eq.~\ref{eq:topic}
    \EndFor
    \EndFor
    \end{algorithmic}
  \end{algorithm}

The overall training objective measures how well we can predict surrounding
words, taking into account the topic distributions of the documents. For each document $d$, given a sequence of words $w_1, w_2,
\cdots, w_{T_d}$, the log-likelihood $\mathcal{L}_d$ is defined as follows.
\begin{equation}
  \small
  \label{eq:hobj}
  \mathcal{L}_d = \sum_{t=1}^{T_d}\sum_{\substack{-c\leq j \leq
      c\\j\neq 0}}\log p(w_{t+j}|w_{t}, d)
  \normalsize
\end{equation}
  where $c$ is the size of the training windows. The overall log-likelihood is then given by $\mathcal{L} = \sum_d \mathcal{L}_d$. 


In the E-step, the topic distribution for each skip-gram in $d$ can be evaluated using the Bayes rule as:
\begin{equation}
  \small
  \label{eq:post}
p(z_{k}^\prime |d, w_t, w_{t+j}) = \frac{p(w_{t+j}|z_{k}^\prime,
  w_t)p(z_{k}^\prime|d)}{\sum_{z} p(w_{t+j}|z, w_{j})p(z|d)}
\normalsize
\end{equation}
In the M-step, given the posterior topic distribution Eq.~\ref{eq:post}, the goal is to maximize the following Q function:
\begin{equation}
  \small
\begin{split}
\mathbf{Q} &= \sum_{d}\sum_{t=1}^{T_d}\sum_{\substack{-c \leq j \leq c\\j\neq 0}}\sum_{z} p(z|d, w_t, w_{t+j})log(p(z|d)p(w_{t+j}|z, w_t))\\
&=\sum_{d}\sum_{\{w_t, w_{t+j}\} \in P_d} n(d, w_t, w_{t+j})\sum_{z} p(z|d, w_t, w_{t+j})\\
&[\log(z|d)+\log(p(w_{t+j}|z,w_t))]
\end{split}
\normalsize
\end{equation}
where $P_d$ is the set of the skip-grams in $d$. $n(d, w_t, w_{t+j})$ denotes the number of the skip-gram $\langle w_t, w_{t+j} \rangle$ in $d$.
Using the Lagrange multiplier, we can obtain the update rule of $p(z|d)$, satisfying the normalization constrains that $\sum_{z} p(z|d)=1$ for each document $d$:
\begin{equation}
  \small
  \label{eq:topic}
p(z|d)=\frac{\sum_{\{w_t, w_{t+j}\}\in
    P_d}n(d,w_t,w_{t+j})p(z|d,w_t,w_{t+j})}{\sum_{\{w_t, w_{t+j}\}\in
    P_d}n(d, w_t, w_{t+j})}
\normalsize
\end{equation}

As already mentioned, it is not feasible to directly optimize $U_{w,z}$ and $V_{w,z}$ due to the term $\sum_{w\in \Lambda}exp(V_{w,z}\cdot
U_{w,z})$. Inspired by the negative sampling scheme, we therefore estimate the probability of predicting the context word $p(w_{t+j}|w_t,z)$ as follows:
\begin{equation}
  \small
  \label{eq:sim}
\begin{split}
\log p(w_{t+j}|w_t, z) & \propto \log \sigma(V_{w_{t+j},z}\cdot U_{w_t,z})\\
&+ \sum_{i=1}^{n}\mathbb{E}_{w_i\sim P}[\log \sigma(-V_{w_i,z}\cdot U_{w_t,z})]
\end{split}
\normalsize
\end{equation}
where $\sigma(x)=1/(1+\exp(-x))$ and $w_i$ is a negative instance
which is sampled from the distribution $P(.)$. Mikolov et al.~\cite{mikolov2013distributed} have investigated many choices for $P(w)$ and found that the best $P(w)$ is equal to the unigram distribution $\textit{Unigram}(w)$ raised to the $3/4rd$ power. We exploit the same setting of $P(w)$ in~\cite{mikolov2013distributed}.
Evaluating $\log p(w_{t+j}|w_t,z)$ for each term in the overall objective function, we obtain the following gradients:
Therefore, the gradients of the objective function with
respect to $U$ and $V$ can be formulated as follows:
\begin{equation}
  \small
  \label{eq:embedding1}
\frac{\partial \mathcal{L}}{U_{w_t,z}} = -(\xi_{w^{\prime} w_t} -
\sigma(V_{w^{\prime},z}\cdot U_{w_t,z})) \cdot V_{w^{\prime},z} \cdot
P(z|d, w_t, w^{\prime})
\normalsize
\end{equation}
\begin{equation}
  \small
  \label{eq:embedding2}
\frac{\partial \mathcal{L}}{V_{w^{\prime},z}} = -(\xi_{w^{\prime} w_t}
- \sigma(V_{w^{\prime},z}\cdot U_{w_t,z})) \cdot U_{w_t,z} \cdot
P(z|d, w_t, w^{\prime})
\normalsize
\end{equation}
where
\begin{equation}
  \small
  \begin{split}
    \xi_{w^{\prime} w_t} =
    \begin{cases}
      1, & \text{if } w^\prime \text{is a word in the context window of $w_t$} \\
      0, & \text{otherwise}\\
      \end{cases}
    \end{split}
    \normalsize
  \end{equation}

  The difference between the updated rules of $U$ and $V$ and those in the original Skip-gram
model is that we maximize $P(z_k|d,w_t, w_{t+j}) \cdot \allowbreak \log P(w_{t+j}|w_t, z)$ instead of $\log P(w_{t+j}|w_t)$ for each skip-gram.
This is in accordance with the fact that our model uses the topic distribution to predict context words. 

\begin{equation}
  \small
    \label{eq:bi-gram}
    \mathcal{L}_{BTM} = \sum_{d}\sum_{t=1}^{T_d}\log p(w_{t+1}|w_{t},
    d)
    \normalsize
  \end{equation}
Comparing the original likelihood of Bi-gram Topic Model (BTM)~\cite{wallach2006topic} in Eq.~\ref{eq:bi-gram} with ours, we can see the connection between our STE model and BTM. Specifically the objective functions in Eq.~\ref{eq:hobj} and Eq.~\ref{eq:bi-gram} share similar form. Both of them are related to the product of conditional probabilities which predict the next word given the preceding word no matter skip-gram or bi-gram. Such connection provides an insight for our model indicating that it is capable of discovering good topics and identifying high-quality word embedding vectors jointly.

  \subsection{Topic Generation}
  \label{sec:top}
  One important aspect of topic models is their interpretability, where the semantic meaning of a topic can be naturally perceived by examining the top ranked words. In standard topic models, these top-ranked words are simply those that maximize $P(w|z)$ for the multinomial distribution associated with the considered topic $z$. In our model, on the other hand, we can evaluate the probability of $p(w_{t+j}|z, w_t)$ for each skip-gram $\langle w_t, w_{t+j} \rangle$. Therefore, we represent each topic as the ranked list of bi-grams. Each bi-gram is sorted using Eq.~\ref{eq:sim} and the top-ranked bi-grams are selected from the ranking list. The
  original time complexity of calculating $p(w_{t+1}|z, w_t)$ is
  $|\Lambda|^2\times K$, where $|\Lambda|$ is the size of the
  vocabulary, i.e., around $10^5$. To make it more efficient, we first
  collect all the bi-grams in the corpus and evaluate the
  corresponding probability $p(w_{t+1}|z, w_t)$. Then the time
  complexity is reduced to be linear to the number of bi-grams. Note that in Eq.~\ref{eq:sim}, we do not need to consider the part related to the sampled negative instances for each bi-gram, i.e., the summation expression, as we can assume it to be constant.

  \subsection{Folding-in for New Documents} 
  Given a new document $d^\prime$, our algorithm can infer the
  topic distribution of $d^\prime$. Given $U$ and $V$ learned from the
  training process, we fix the values of $U$ and $V$, and then only
  update $p(z|d^\prime)$ using Algorithm~\ref{alg:em}.
  
  For each word $w$ in $d^\prime$, the posterior topic distribution of
  $w$, $p(z|w, d^\prime)$
  can also be inferred. We consider that the topic distribution of $w$
  is related to not only its context words but also the topic
  distribution of $d^\prime$. Therefore, using the Bayes rule, we have:
  \begin{equation}
    \small
  \label{eq:inf}
    \log p(z|w, c_w, d^\prime) \propto \log p(z|d^\prime) + \log p(c_w
    | z, w, d^\prime)
    \normalsize
  \end{equation}
  where $c_w$ is the set of the context words of $w$. The likelihood term $\log p(c_w| z, w, d^\prime)$ can be defined as the sum of
  $\log p(w_{t+j}|w_{t}, z)$, where $w_{t+j}$ belongs to the context
  words.
  The probability $p(w_{t+j}|w_t, z)$ can be computed in Eq.~\ref{eq:sim}. The term $p(z|d^\prime)$ is the corresponding prior probability.
  %

\section{Experiments}

In this section, we present a detailed analysis of the performance of our method. We first present a qualitative analysis of the learned topic-specific embeddings. We then focus on evaluating the quality of the word embedding on a standard word similarity benchmark. Subsequently, we evaluate the quality of the identified topics, focusing in particular on their coherence. Finally, as an example of a downstream task, we analyze the performance of our model in a document classification task.


\subsection{Qualitative Analysis}
\label{sec:word_sim}

\begin{figure*}[!h]
  \begin{subfigure}{0.7\textwidth}
    \centering
    \includegraphics[width=0.99\linewidth]{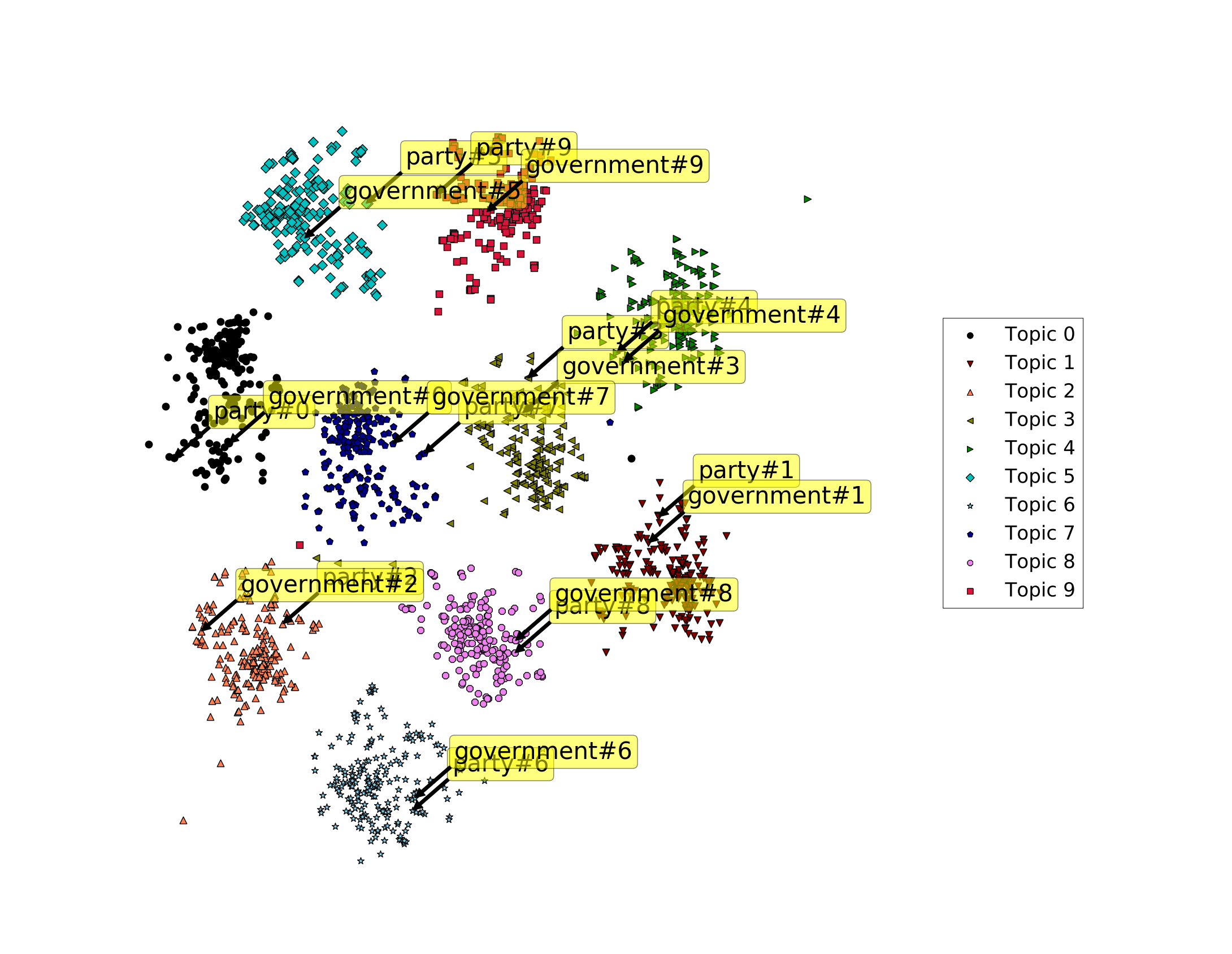}
    \caption{STE-Same}
    \label{fig:ste-same}
  \end{subfigure}
  \begin{subfigure}{0.7\textwidth}
    \centering
    \includegraphics[width=0.99\linewidth]{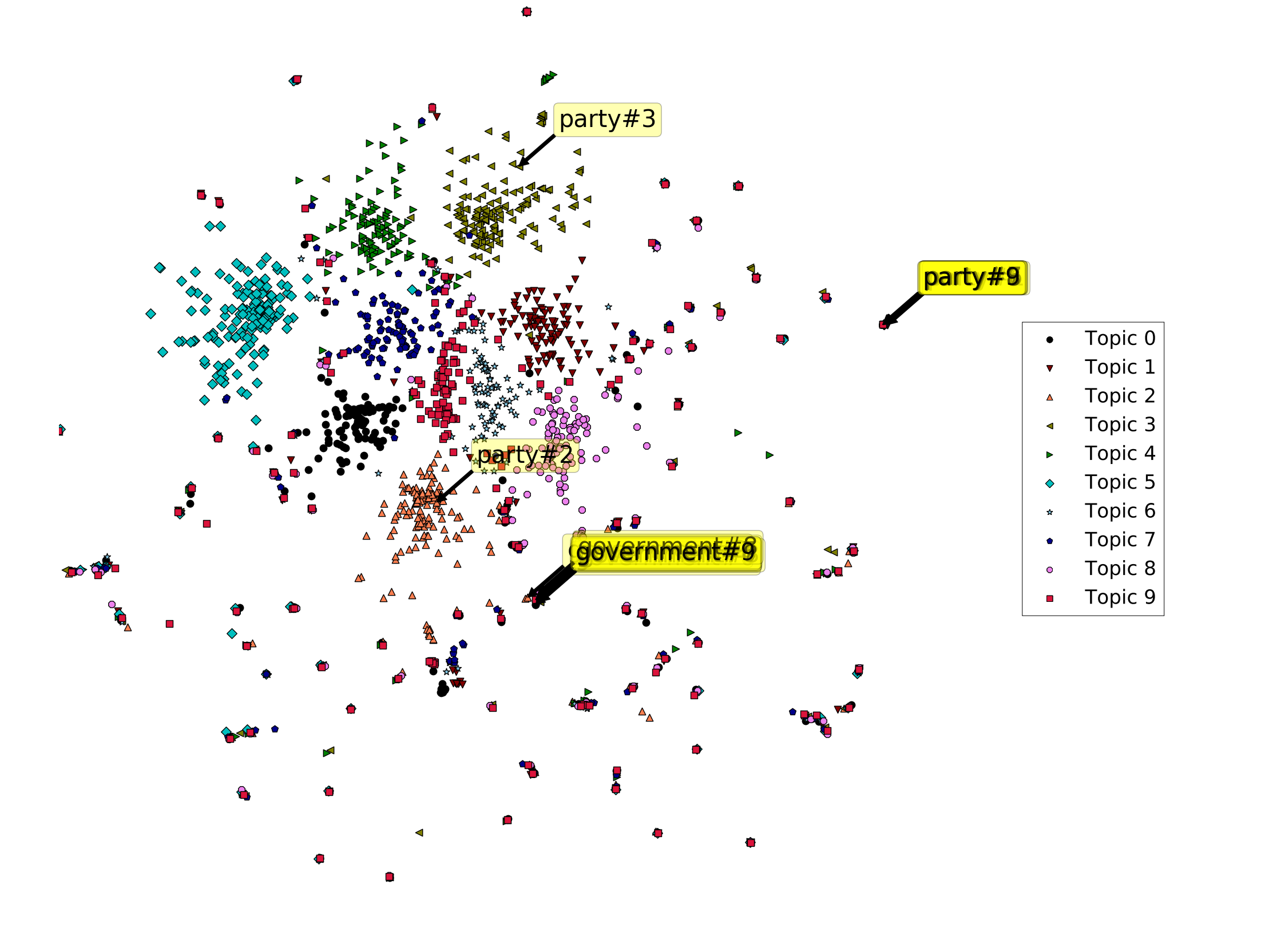}
    \caption{STE-Diff}
    \label{fig:ste-diff}
  \end{subfigure}
  \caption{Visualization of the word embeddings learned using STE-Same and STE-Diff with 10 topics. The polysemous word ``party'' and the monosemous word ``government'' are highlighted for the comparison.\label{figSTE}}
  \label{fig:ste}
\end{figure*}

To illustrate the interactions between word embeddings and latent topics, we visualize the results of our STE-Same and
STE-Diff models in Figures \ref{fig:ste-same} and \ref{fig:ste-diff} respectively. For this figure, and in the following experiments, we have used the Wikipedia dump from April 2010 \cite{shaoul2010westbury}, which has previously been used for other word embedding models~\cite{huang2012improving}. We have chosen the number of topics $K = 10$. The number of outer iterations and inner iterations are both set to 15. The dimension of the embedding vectors was chosen as 400, in accordance with~\cite{liu2015topical}. For each skip-gram, we set the window size to 10 and sample 8 negative instances following~\cite{tian2014probabilistic}. 
To generate the visualization in Figure \ref{figSTE}, we have used the t-SNE algorithm~\cite{maaten2008visualizing}, applied to the vectors of the 500 most frequent words.



In Figure~\ref{fig:ste}, each node denotes a topic-specific word vector. To illustrate how polysemy is handled, we show labels for the word ``party'', as an example of a polysemous word, and for the word ``government'', as an example of a monosemous word. The labels show both the word and topic index, separated by ``\#''. In Figure~\ref{fig:ste-same}, we can observe that our STE-Same model divides
the whole space into K disjoint subspaces, with each subspace representing a topic. Within each subspace the words with similar meanings are close. Note that the similarity of the words ``government'' and ``party'' depend on the considered sense for the latter word. Accordingly, we see that ``government'' and ``party'' are close to each other in some subspaces, but far apart in others. For example, in the
subspace of Topic 0, the position of the word ``party'' is far from
the position of the word ``government'', which suggests that the meaning of ``party'' under Topic 0 is not related to a political party. In contrast, for Topics 4, 6 and 8, the vectors for ``party'' and  ``government'' are similar, suggesting that ``party'' in these spaces is regarded as a political organization.


On the other hand, Figure~\ref{fig:ste-diff} illustrates how STE-Diff generates a more universal space in which word embeddings from different topics co-exist in this shared space. Words from different topics with a similar meaning are represented using similar vectors. In particular, for monosemous words such as ``government'' the word vectors are approximately the same. For the word ``party'', on the other hand, we see three clearly distinct representations, only one of which (party\#2) is close to the vectors for ``government''. Moreover, we found that ``party\#3'' represents the semantic sense of community because it is close to the word ``organization'' and the word ``group''. The representations of the word ``party'' from the other topics are approximately the same. They are close to the representations of the word ``summer'' and the word ``shout''. It indicates that the word ``party'' represents the meaning about the concept of human activity.

From the comparison between Figure~\ref{fig:ste-same} and Figure~\ref{fig:ste-diff}, the STE-Same model and the STE-Diff model can be regarded as two different paradigms derived from the treatment of topic consistency in a skip-gram. The advantage of the STE-Diff model over the STE-Same model is that the STE-Diff model can support better the evaluation of the similarity of words from different topics. For example, the senses of ``party'' under Topics 4 and 6 are very close to ``government'' in the STE-Same model. However, when we evaluate the similarity between ``party\#4'' and ``government\#6'', we find that the distance cannot reflect very well the word similarity. Nevertheless, this STE-Same model can still achieve comparable performance with existing models in the quantitative word embedding evaluation experiment as presented in the next subsection. On the other hand, our STE-Diff model can handle very well the evaluation of the similarity of words from different topics. The reason is that it represents every sense in a more universal shared space without gaps between different topics.

\begin{table}[!h]
  \caption{The most similar words identified by the original Skip-gram
    model and our STE-Diff model.}
  \label{table:similar_word}
  \small
  \begin{tabular}{|c|c|c|}
    \hline
    \textbf{Model} & \textbf{Words} & \textbf{Similar Words} \\
    \hline
    Skip-gram & apple & macintosh, ios, juice\\
    \hline
    \multirow{2}{*}{STE-Diff} & apple\#1 & peach, orange, juice\\
    & apple\#2 & macintosh, ipod, windows\\
    \hline
    Skip-gram & java & sumatra, html, somalia\\
    \hline
    \multirow{2}{*}{STE-Diff} & java\#1 & sumatra, somalia, sudan\\
    & java\#2 & html, lisp, jde\\
    \hline
    Skip-gram & cell & phones, viral, biology\\
    \hline
    \multirow{2}{*}{STE-Diff} & cell\#1 & phones, technology, scanner\\
    & cell\#2 & viral, tumor, embryonic\\
    \hline
  \end{tabular}
  \normalsize
  \end{table}
Table~\ref{table:similar_word} shows the nearest neighbours of
some polysemous words, according to the Skip-gram model and our STE-Diff model (using cosine similarity in both cases). We observe that these nearest neighbours for Skip-gram mix different senses of the given words, which is expected since Skip-gram does not address polysemy. For example, the nearest neighbours of ``apple'' are given as ``macintosh'', ``ios'', and ``juice'', indicating the ``company'' and ``fruit'' interpretations of the word ``apple'' are mixed. In contrast, our STE-Diff model can distinguish different prototypes of polysemous words via the latent topics. For example, the most similar words of ``apple'' under Topic 1 are ``peach'', ``orange'' and ``juice'', which clearly corresponds to the fruit interpretation. Under Topic 2, they are ``macintosh'', ``ipod'' and ``windows'', clearly referring to the company interpretation. 


\subsection{Word Embedding Evaluation}
The most common approach for evaluating word embeddings is to assess how well the similarity scores they produce correlate with human judgments of similarity. Although there are several word similarity benchmark datasets, most do not provide any context information for the words, and are therefore not appropriate for evaluating models of similarity for polysemous words. 
Huang et al.~\cite{huang2012improving} prepared a data set, named
Stanford's Contextual Word Similarity (SCWS) data set, which
includes 2003 word pairs together with their context sentences. The ground truth similarity score with the range $[0, 10]$ was labeled by humans, according
to the semantic meaning of the words in the given contexts. We adopt this benchmark data set for evaluating the quality of our word embeddings.


We compare our results with the following baselines and state-of-the-art methods, reporting the previously published results from their papers.

\begin{description}
  \item[TFIDF] We consider two variants, TFIDF and Pruned TFIDF. The TFIDF method represents each word as a vector, capturing the context words with which it co-occurs in a 10-word window. Each context word is presented by the one-hot representation and weighted via TF-IDF learned from the training set. The Pruned TFIDF method
    proposed by~\cite{reisinger2010multi} improves the original TFIDF model by pruning the context words with low TF-IDF scores.
  \item[Word embedding] These baselines include the C\&W method proposed
    by~\cite{collobert2008unified} and the Skip-gram
    model~\cite{mikolov2013distributed}. Note that since neither of these methods considers polysemy, word similarity is evaluated without regarding the context sentences.  
  \item[Topic models] The first model, named
    LDA-S, represents each word $w$ in a document $d$ as the posterior topic distribution, namely, $p(z|w)$ where $p(z|w)\propto p(w|z)p(z|d)$. The second model, named LDA-C, additionally considers the
    posterior topic distribution of the surrounding context words, as follows:
    \begin{equation}
      p(z|w,c) \propto p(w|z)p(z|d) \prod_{w^{\prime} \in c} p(w^{\prime}|z)
      \end{equation}
      where $c$ is the set of the context words of $w$.
  \item[Multiple prototype models] These methods represent each
    word sense as a fixed length vector. One representative work,
    proposed by Huang et al.~\cite{huang2012improving}, exploits global
    properties of the corpus such as term frequency to learn multiple
    embeddings via neural networks. Tian et al.~\cite{tian2014probabilistic} introduce a latent variable to denote
the distribution of multiple prototypes for each word in a
probabilistic manner. Liu et al.~\cite{liu2015topical} propose a model
called TWE, which concatenates the pre-trained topics with the
word embeddings for representing each prototype.
  \end{description}

\noindent We present the related parameter settings as reported in the previous papers~\cite{huang2012improving, tian2014probabilistic, liu2015topical}. For Pruned TFIDF, top 200 words with the highest scores are preserved. For the model in~\cite{tian2014probabilistic}, each word is assumed with 10 prototypes. Following~\cite{liu2015topical}, the number of topics of LDA-S, LDA-C and TWE is 400. Note that the size of each embedding vector in TWE is 800~\cite{liu2015topical}, which consists of 400-dimension word embedding and 400-dimension topic embedding. The parameter setting of our STE model is the same as described in Section~\ref{sec:word_sim}.

For all the different representations, the word similarity is evaluated using cosine similarity. However, for the multiple prototype based methods, as well as for our STE model, two different variants are considered: 

\begin{description}
\item[\textit{AvgSimC}] Given a word $w$ and its associated context words $c_w$, we can infer the posterior topic distribution $p(z| w, c_w, d)$ according to Eq.~\ref{eq:inf}. Then the averaged similarity between two words ($w_i, w_j$) over the assignments of topics is computed as:
\begin{align*}
  \quad\quad\quad\textit{AvgSimC}(w_i, w_j) = \sum_{z_i} \sum_{z_j} &p(z_i|w_i,c_{w_i}) p(z_j|w_j, c_{w_j})\\ 
  &\times \cos(U_{w_i, z_i}, U_{w_j, z_j})  
\end{align*}
where $U(w_i, z_i)$ is the embedding vector of $w_i$ under the topic $z_i$ and  $\cos(\cdot)$ is the cosine similarity. 
\item[\textit{MaxSimC}] In this case, we instead evaluate the similarity between the most
  probable vectors of each word. It is defined as:
  \begin{equation*}
      \textit{MaxSimC}(w_i, w_j) = Sim(U_{w_i, z_i}, U_{w_j, z_j})
    \end{equation*}
    where $z = \arg\max_{z}(p(z|w, c))$.
\end{description}


Following previous work, we use the Spearman correlation
coefficient as the evaluation metric. The results are shown in
Table~\ref{table:sim_res}. The STE model performs comparably to the state-of-the-art TWE model, and outperforms the baseline methods. TWE also exploits both topics and embeddings. However, the vectors in the TWE model have twice as many dimensions as those in our model, since each word is represented as the concatenation of a word vector and a topic vector.
The STE-Diff variant outperforms STE-Same,
for both the $AvgSimC$ and $MaxSimC$ ranking measures. While STE-Same can generate more coherent topics as indicated in Section~\ref{sec:coherence}, this comes at the price of slightly less accurate word vectors, which is not unexpected.


It is interesting to note that the original Skip-gram model can still achieve satisfactory performance. We observe that the words in the SCWS data set are mostly monosemous words. Particularly, among 2003 word pairs in this data set, there are 241 word pairs containing the same word within a word pair. One may expect that such identical words have different senses leading to a low ground truth similarity score. However, only 50 of them have the ground truth similarity score less than 5.0. It indicates that the proportion of the challenging polysemous words in this data set is quite small.


\begin{table}[!ht]
\caption{Spearman correlation $\rho \times 100$ for the SCWS data set.}
  \label{table:sim_res}
  \centering
  \small
\begin{tabular}{|c|c|c|}
  \hline
  \textbf{Model} & \textbf{Similarity Metrics} & $\rho \times 100$ \\
  \hline
  C\&W & Cosine Similarity & 57.0 \\
  Skip-gram & Cosine Similarity & 65.7 \\
  TFIDF & Cosine Similarity & 26.3 \\
  Pruned TFIDF & Cosine Similarity & 62.5 \\
  LDA-S & Cosine Similarity & 56.9 \\
  LDA-C & Cosine Similarity & 50.4  \\
  \hline
  Tian & $AvgSimC$ & 65.4 \\
  Tian & $MaxSimC$ & 63.6 \\
  Huang & $AvgSimC$ & 65.3 \\
  Huang & $AvgSimC$ & 58.6 \\
  TWE & $AvgSimC$ & 68.1 \\
  TWE & $MaxSimC$ & 67.3 \\
  STE-Same & $AvgSimC$ & 66.7 \\
  STE-Same & $MaxSimC$ & 65.5 \\
  STE-Diff & $AvgSimC$ & 68.0 \\
  STE-Diff & $MaxSimC$ & 67.7 \\
  \hline
\end{tabular}
\normalsize
\end{table}




\subsection{Topic Coherence}
\label{sec:coherence}
In our model, topics can be interpreted by looking at the top-ranked bi-grams according to 
Eq.~\ref{eq:sim}. To evaluate how coherent these topics are, we have applied our model to the training set of the 20Newsgroups corpus. The 20Newsgroups corpus\footnote{http://qwone.com/~jason/20Newsgroups/} is a collection of 19,997 newsgroup
documents. The documents are sorted by date and split into training set ($60\%$) and test
set ($40\%$). The data is organized, almost evenly, into 20 different newsgroups, each
corresponding to a different topic. Some of the newsgroups are very closely related to each other, however. The categories of this corpus, partitioned according to subject matter. are shown in Table~\ref{table:category}. As text preprocessing, we have removed punctuations and stop words, and all words were lowercased.


\begin{table}[h]
  \caption{The categories of the 20Newsgroups corpus.}
  \small
  \label{table:category}
  \begin{tabular}{|@{\hspace{4pt}}l@{\hspace{2pt}}|@{\hspace{4pt}}l@{\hspace{2pt}}|@{\hspace{4pt}}l@{\hspace{2pt}}|}
    \hline
    \tabincell{l}{comp.graphics \\
comp.os.ms-windows.misc \\
comp.sys.ibm.pc.hardware \\
comp.sys.mac.hardware \\
comp.windows.x \\
    } & \tabincell{l}{
        rec.autos \\
rec.motorcycles \\
rec.sport.baseball \\
rec.sport.hockey } & \tabincell{l}{sci.crypt \\
sci.electronics \\
sci.med \\
    sci.space \\} \\
    \hline
misc.forsale & \tabincell{l}{talk.politics.misc \\
talk.politics.guns \\
talk.politics.mideast	\\talk.religion.misc\\} & \tabincell{l}{
alt.atheism \\
    soc.religion.christian \\} \\
    \hline
  \end{tabular}
  \normalsize
\end{table}

To evaluate the generated topics, we use a common topic coherence metric which measures the
relatedness between the top-ranked words \cite{chang2009reading,stevens2012exploring}. The intuition is that topics where the top-ranked words are all closely semantically related are easy to interpret, and in this sense semantically coherent. Following Lau et
al.~\cite{newman2010automatic, lau2014machine}, we use the pointwise mutual information (PMI) score as our topic coherence metric. PMI has been found to strongly
correlate with human annotations of topic
coherence. For a topic $z$, given the top-ranked $T$ words, namely,
$w_1, w_2,\cdots, w_T$, the PMI score of the topic $z$ can be calculated as follows:
\begin{equation}
  \small
  \label{eq:PMI}
\text{PMI-Score}(z) = \sum_{1\leq i\leq j \leq
  T}\log\frac{P(w_i, w_j)}{P(w_i)P(w_j)}
\normalsize
\end{equation}
where $P(w_i, w_j)$ represents the probability that words $w_i$ and $w_j$ co-occur and $P(w_i) = \sum_w P(w_i,w)$. 
We compute the average PMI
score of word pairs in each topic using Eq.~\ref{eq:PMI}. The average
score of all the topics is computed. The higher the value, the better is the coherence. Newman et
al. \cite{newman2010automatic} observed that it is important to use another data set to evaluate the PMI based measure. Therefore, we use a 10-word sliding window in Wikipedia~\cite{shaoul2010westbury} to estimate the probabilities $P(w_i,w_j)$.

We compare our STE model with the Bi-gram Topic Model
(BTM)~\cite{wallach2006topic} which predicts the next word given the
topic assignments as well as the current word. Note that our STE model and LDA are not
directly comparable because LDA can only output unigrams as topics.



All models are trained on the training set of the 20newsgroups
corpus. The number of topics was set to 20 for all models, which is
the same as the number of categories in the 20Newsgroups corpus. For
our model, 400-dimensional word vectors were used. To extract the
top-ranked words, we first learn the topic-specific word embeddings
$U_w$ and $V_w$ and then use Eq.~\ref{eq:sim}. Only bi-grams with
frequency greater than 5 are considered. For the BTM model, we use the default setting of hyper-parameters provided by the package~\footnote{http://mallet.cs.umass.edu/topics.php}, i.e., $\alpha = 50.0$, $ \beta = 0.01$, and
$\gamma=0.01$.



\begin{table}[t]
\caption{Topic coherence evaluation using the PMI metric with
  different numbers of top words.}
\small
\begin{center}
\begin{tabular}{|c|c|c|c|c|}
  \hline
  & $T = 5$ & T = 10 & T = 15 & T = 20 \\
  \hline
  \textbf{BTM} & 0.014 & 0.036 & 0.041 & 0.048\\ 
  \hline
  \textbf{STE-Same} & 0.180 & 0.110 & 0.107 & 0.102\\
  \hline
  \textbf{STE-Diff} & 0.015 & 0.067 & 0.058 & 0.054\\
  \hline
\end{tabular}
\end{center}
\normalsize
\vspace{-1.2em}
\label{table:coherence_wiki}
\end{table}

The average PMI scores with different numbers of top-ranked words are shown in Table~\ref{table:coherence_wiki}. We
can see that our STE model generally improves the coherence of the
learned topics. Compared with BTM, our STE model incorporates the semantic relationship between words which is learned
from word embeddings to improve the quality of topics.
Compared with the variant STE-Diff, the STE-Same model can produce more coherent topics as presented in Figure~\ref{fig:ste}.


Table~\ref{table:top_words} shows the top 10 bi-grams of the 
topics identified by the STE-Same model. We can observe that many topics are corresponding to categories from
Table~\ref{table:category}. For example, the top-ranked bi-grams of Topic 1, such
as the terminology phrase ``remote sensing'' and the
name of the company, ``mcdonnell douglas'', indicate that Topic 1 is
related to the category ``sci.space''. Similarly, most of the top-ranked
bi-grams of Topic 5 are medical terms, such as ``mucus membrane'' and ``amino acids''. The top-ranked bi-grams of
Topic 6 are the names of some famous baseball and hockey players, such
as ``brind amour''\footnote{https://en.wikipedia.org/wiki/Rod\_Brind'Amour} and ``garry galley''\footnote{https://en.wikipedia.org/wiki/Garry\_Galley}, indicating that Topic 6 is associated to the categories
``rec.sport.baseball'' and ``rec.sport.hockey''. Among others, we also observe a connection between Topic 4 and ``talk.politics'', Topic 7 and ``comp.sys.ibm.pc.hardware'', as well as Topic 8 and ``soc.religion.christian''. These
correspondences further illustrate that our model can identify latent topics effectively.


\begin{table*}[!h]
  \centering
  \caption{Top 10 bi-grams of the generated topics from the 20Newsgroups
    data set.}
  \label{table:top_words}
  \small
  \begin{tabular}{|c|c|c|c|c|c|c|c|c|c|}
    \hline
    \textbf{Topic 1} & \textbf{Topic 2} & \textbf{Topic 3} & \textbf{Topic 4} \\
    \hline
mcdonnell douglas & graphlib inria & carpal tunnel & deir yassin \\
remote sensing & inria graphlib & blaine gardner & ottoman empire \\
southern hemisphere & rainer klute & vernor vinge & bedouin negev \\
northern hemisphere & burkhard neidecker & thermal recalibration & negev
                                                                 bedouin \\
ozone layer & renderman pixar & syndrome carpal & ermeni mezalimi \\
cosmic rays & pixar renderman & tunnel syndrome & ishtar easter \\
orthodox physicists & spine dinks & mudder disciples & eighth
                                                         graders \\
black holes & reflections shadows & lance hartmann & prime minister
                                                        \\
ames dryden & marching cubes & hartmann lance & nagorno karabakh \\
sounding rockets & liquid fueled & blgardne javelin & lieutenant
                                                         colonel \\
circular orbit & gravity assist & evans sutherland & democracy corps\\
    \hline
    \textbf{Topic 5} & \textbf{Topic 6} & \textbf{Topic 7} & \textbf{Topic 8} \\
    \hline
mucus membrane & spring training & setjmp longjmp & saint aloysius \\
amino acids & brett hull & xtaddtimeout xtaddworkproc & empty tomb \\
kidney stones & cape breton & xtaddinput xtaddtimeout & respiratory
                                                         papillomatosis \\
anti fungal & garry galley & bname pname & zaurak kamsarakan \\
candida albicans & brind amour & babak sehari & biblical contradictions \\
oxalic acid & bobby clarke  & infoname uuname & recurrent
                                                  respiratory \\
kirlian photography & cheap shot & uuname infoname &ohanus
                                                       appressian \\
kidney stone & wade boggs & physicist dewey & archbishop lefebvre\\
glide jubilee & wayne komets & richardd hoskyns &joseph smith \\
penev venezia & tommy soderstrom & arithmetic coding &serdar argic \\
vaginal tract & chris chelios & mcdonnell douglas & rodney king \\
    \hline
    
    \end{tabular}
    \normalsize
\end{table*}

Another interesting observation is that most of top-ranked bi-grams are person names and domain-specific terms. This arises because we rank
the bi-grams according to the probability $p(w_{i+1}|w_{i}, z)$, which
indicates that the words $w_{i+1}$ and $w_i$ should have strong connections
under the topic $z$. The name of famous people and
domain-specific terms can satisfy this requirement.



\subsection{Document Classification}
We also analyze the suitability of our model for representing documents. To this end, we consider the task of document classification,  using again the 20Newsgroup corpus. We have used the original splits into training ($60\%$) and testing
 ($40\%$) data. For our STE model, as before, we set the number of topics as 20 and the dimensionality of the vector space as 400. To apply our model to document representation, we first infer the posterior topic distribution of each word in the test set using
Eq.~\ref{eq:inf}, and then represent each document as the average of the word vectors in the
document, weighted by the TF-IDF score of each word. In particular, each word vector is given by:
\begin{equation}
  \text{Vec}_w = TFIDF_w \times \sum_{z=1}^K p(z|w, c)U_{w,z}
  \end{equation}
where $TFIDF_w$ is the TF-IDF score of $w$ in the document. To perform document classification, we use a linear
SVM classifier using the package from~\cite{fan2008liblinear}.


We compare our approach with several other methods for representing documents, including
bag-of-words, vector-space embeddings and latent topics. BOW is the standard
bag-of-words model, which represents each document by weighting terms using the TFIDF
score. As a form of feature selection, we only consider the 50,000 most frequent words in the bag-of-words representation. The embedding-based methods
include the word embedding method Skip-gram, a document embedding method called the Paragraph
Vector model (PV)~\cite{le2014distributed}, and the TWE model which also considers topics. For the Skip-gram model, we set the number of dimension to 400. We represent each document as the average of word vectors weighted by TFIDF scores. 
The PV model proposed by~\cite{le2014distributed} represents
each document directly as a vector. We use the doc2vec implementation~\footnote{http://github.com/ccri/gensim} for PV. For
 TWE, we report the experimental results published
in~\cite{liu2015topical}. The topic based methods include LDA,
LFTM~\cite{nguyen2015improving}, and GPU-DMM~\cite{li2016topic}. These models represent each document via the
posterior topic distributions. For LFTM, we reported the experimental result published in~\cite{nguyen2015improving}. Only F-measure of LFTM is provided. The number of topics for LFTM is 80, as that value was reported to lead to the best performance in~\cite{nguyen2015improving}. The GPU-DMM model promotes semantically related words learned from pre-trained word embeddings by using the generalized Polya urn model. We use the default parameter setting in the GPU-DMM model. We use a number of standard evaluation metrics for classification tasks, namely accuracy, precision, recall and F1 measure. 

\begin{table}[!h]
  \centering
  \caption{Document classification results.}
  \label{table:classification}
  \begin{tabular}{|c|c|c|c|c|}
    \hline
    \textbf{Model} & \textbf{Accuracy} & \textbf{Precision} & \textbf{Recall} & \textbf{F-measure} \\
    \hline
    BOW & 79.7 & 79.5 & 79.0 & 79.0 \\
    Skip-Gram & 75.4 & 75.1 & 74.3 & 74.2 \\
    TWE & 81.5 & 81.2 & 80.6 & 80.6 \\
    PV & 75.4 & 74.9 & 74.3 & 74.3 \\
    LDA & 72.2 & 70.8 & 70.7 & 70.0 \\
    LFTM & - & - & - & 76.8 \\
    GPU-DMM & 48.0 & 56.8 & 46.9 & 47.3 \\
    STE-Same & 80.4 & 80.3 & 80.4 & 80.2 \\
    STE-Diff & \textbf{82.9} & \textbf{82.5} & \textbf{82.3} & \textbf{82.5} \\
    \hline
  \end{tabular}
  \vspace{-2em}
  \end{table}
  The results of the document classification task are presented in
  Table \ref{table:classification}, showing that our STE-Diff model achieves the
  best results. Good results are also obtained for the TWE model, which suggests that the quality of document representations can clearly benefit from combining topics models with word embeddings. Nevertheless, compared to TWE, our model can take advantage of the interaction between topics and word embeddings to improve both, and is thus able to outperform the TWE model. The performance of GPU-DMM is not as good as expected. One reason is that this model is proposed for handling short texts~\cite{li2016topic}. Unfortunately, the documents in 20Newsgroups are not short texts.  


\section{Conclusions}
We have proposed a model that jointly learns word embeddings and latent topics. Compared to standard word embedding models, an important advantage of incorporating topics is that polysemous words can be modeled in a more principled manner. Compared to standard topic models, using word embeddings can achieve superiority because more coherent topics can be obtained. While some previous works have already considered combinations of word embedding and topic models, these works have relied on a two-step approach, where either a standard word embedding was used as input to an improved topic model, or a standard topic model was used as input to an improved word embedding. In contrast, we jointly learn both word embeddings and latent topics, allowing our model to better exploit the mutual reinforcement between them. We have conducted a wide range of experiments, which demonstrate the advantages of our approach.


\bibliographystyle{ACM-Reference-Format}
\bibliography{sigproc} 

\end{document}